\documentclass[10pt]{article}

\usepackage[utf8]{inputenc}
\usepackage[T1]{fontenc}
\usepackage{lmodern}
\usepackage[margin=0.85in]{geometry}
\usepackage{amsmath,amssymb,mathtools}
\usepackage{booktabs}
\usepackage{multirow}
\usepackage{array}
\usepackage{adjustbox}
\usepackage{xcolor}
\usepackage{enumitem}
\usepackage{graphicx}
\usepackage{caption}
\usepackage{hyperref}
\usepackage[numbers,sort&compress]{natbib}
\usepackage{listings}
\usepackage{balance}
\usepackage{authblk}

\hypersetup{
  colorlinks=true,
  linkcolor=blue,
  citecolor=blue,
  urlcolor=blue,
  pdftitle={SaiVLA-JEPA: Evaluating Future Latent Prediction in JEPA-Conditioned VLA Control},
  pdfauthor={Rufeng Xiao, Ruihao Xiao, Wei Chen, Xinhai Sun},
  pdfsubject={JEPA-conditioned flow-matching vision-language-action control and test-time adaptation}
}
\setlist{nosep,leftmargin=*}
\newenvironment{reporttable}[1][0.88\linewidth]{%
  \centering\begin{minipage}{#1}\centering\small
  \setlength{\tabcolsep}{5pt}\renewcommand{\arraystretch}{1.12}%
}{\end{minipage}\par}
\newcommand{\reporttablenote}[1]{%
  \par\vspace{5pt}\begin{minipage}{\linewidth}\footnotesize\raggedright #1\end{minipage}%
}

\newcommand{\foursixteen}{FutureJEPA}
\newcommand{\Flow}{\mathcal{L}_{\mathrm{flow}}}
\newcommand{\Lfuture}{\mathcal{L}_{\mathrm{future}}}
\newcommand{\Ltotal}{\mathcal{L}_{\mathrm{total}}}

\title{SaiVLA-JEPA: Evaluating Future Latent Prediction in JEPA-Conditioned VLA Control}

\author{Rufeng Xiao\textsuperscript{1}\quad Ruihao Xiao\textsuperscript{1}\quad Wei Chen\textsuperscript{1}\quad Xinhai Sun\textsuperscript{1}}
\date{}

\begin{document}
\maketitle
\footnotetext[1]{VLA Research Team, Synthoid.ai, Shanghai, China. Email: \texttt{scholar@synthoid.ai}}

\begin{abstract}
Joint-embedding predictive architectures (JEPAs) learn to predict representations rather than pixels, offering a compact way to model how visual scenes evolve.  We ask whether this predictive objective improves closed-loop control when a vision-language-action (VLA) policy already uses V-JEPA features.  We call our JEPA-conditioned flow policy family SaiVLA-JEPA. It combines a vision-language model with a diffusion transformer action expert and receives features from a frozen V-JEPA encoder.  Two lightweight auxiliary heads predict future representations during training and are omitted at inference, leaving the action sampler unchanged.  We compare policies without JEPA features, with current JEPA conditioning, and with additional future supervision on 10{,}030 paired LIBERO-Plus episodes.  Current JEPA conditioning reaches about 82.5\% success versus 71.6\% without JEPA, while future supervision provides no detectable aggregate gain at the evaluated training budget.  We separately evaluate SaiVLA-JEPA-TCA, which adapts latent context, and SaiVLA-JEPA-TPA, which adapts selected parameters. The context-retry gain is a system-level result under a larger attempt budget; parameter adaptation shows no consistent advantage over the frozen policy and target controls. These results motivate assessing latent-prediction objectives through paired closed-loop rollouts rather than latent loss alone.

\textbf{Keywords:} JEPA, V-JEPA, vision-language-action, flow matching, future representation prediction, test-time adaptation, LIBERO-Plus.
\end{abstract}

\section{Introduction}
Vision-language-action policies must preserve semantic information while producing stable, low-latency actions.  We consider a VLM-conditioned flow-matching policy in which a frozen V-JEPA encoder supplies an additional visual representation to a DiT action expert \cite{bardes2024vjepa,mur2026vjepa21,black2024pi0}.  The representation is useful for conditioning, but the action loss alone does not explicitly require the policy trunk to model how the scene representation changes over time.

We call the shared JEPA-conditioned flow-matching policy family \textbf{SaiVLA-JEPA}. Within this family, CurrentJEPA denotes the matched current-feature control and FutureJEPA the variant with future-latent supervision. At test time, we study two update levels: \textbf{SaiVLA-JEPA-TCA} performs test-time context adaptation by inferring persistent and transient latent contexts while keeping network weights frozen, whereas \textbf{SaiVLA-JEPA-TPA} performs test-time parameter adaptation by updating a selected subset of network weights. We use TCA and TPA as short names after this definition; the two routes are evaluated separately rather than as consecutive stages.

\begin{figure}[t]
\centering
\includegraphics[width=\linewidth]{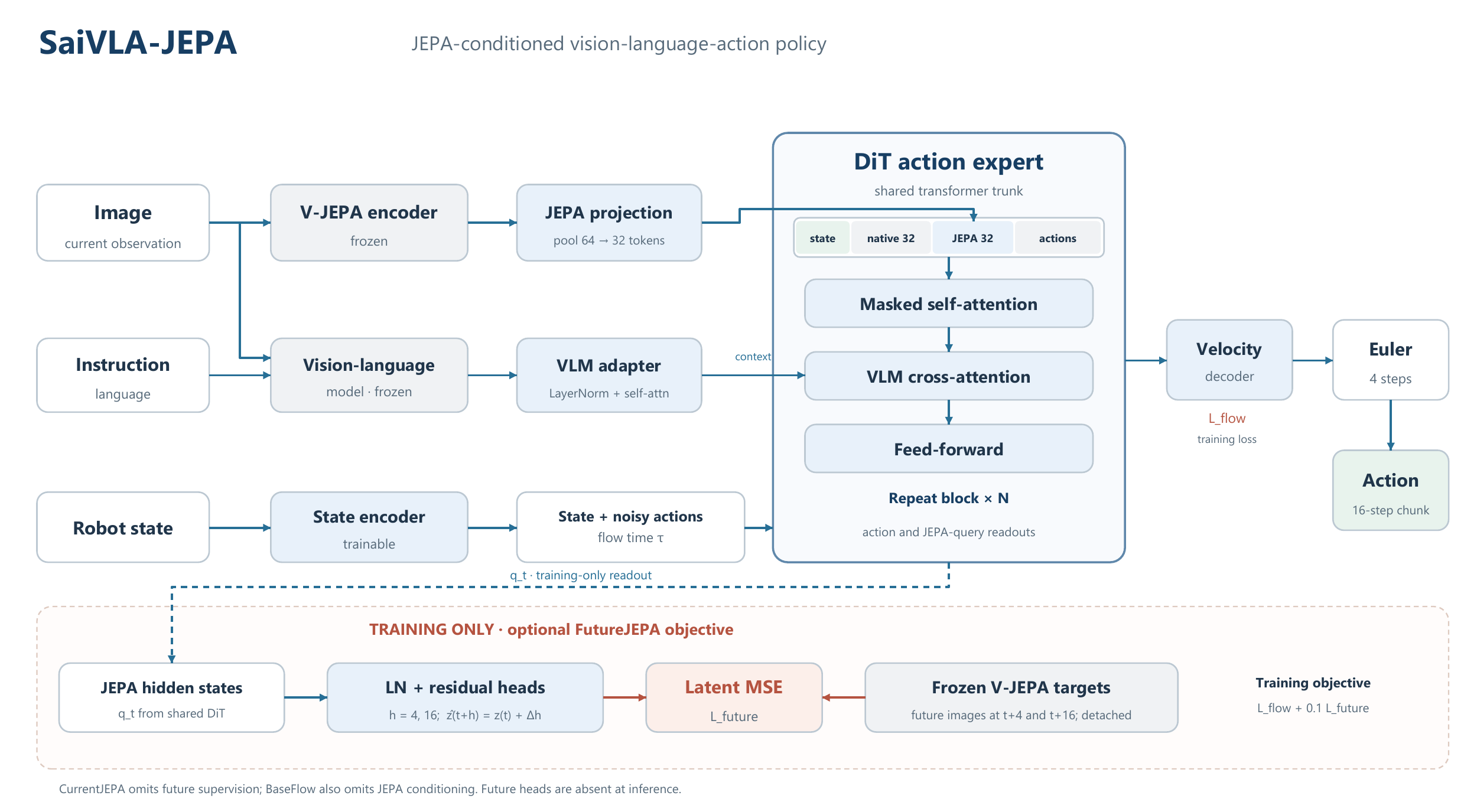}
\caption{SaiVLA-JEPA architecture and its FutureJEPA training variant.
Current images feed the frozen V-JEPA and VLM encoders; projected JEPA tokens
join the native prefix and action tokens in the repeated DiT action expert,
while VLM features enter through per-block cross-attention and robot state
is encoded separately. Four Euler evaluations produce a 16-action chunk.
For FutureJEPA, JEPA-token hidden states from the shared DiT trunk feed residual heads at
offsets 4 and 16, supervised by detached targets from a frozen V-JEPA
encoder. This auxiliary branch is absent at inference. CurrentJEPA omits
future supervision; BaseFlow also omits JEPA conditioning.}
\label{fig:architecture}
\end{figure}

We study a focused intervention: add future latent prediction without changing the action parameterization, the native flow prefix, or the inference-time sampler. Figure~\ref{fig:architecture} summarizes the policy and ablation arms. The treatment predicts latents at horizons \(h\in\{4,16\}\). CurrentJEPA uses current JEPA conditioning and the flow action objective without future supervision; BaseFlow removes JEPA conditioning. All three policies are trained for 25,000 steps and evaluated on the same benchmark cases.

The completed paired comparison finds no detectable aggregate success-rate benefit from future supervision at this training budget. Under the final mixed-cap evaluation, BaseFlow reaches 71.63\% overall and 53.12\% on LIBERO-10, compared with 81.99\% and 73.84\% for FutureJEPA. This is a joint contrast because the policies differ in both JEPA conditioning and future supervision. The future-head estimand and its paired statistics are reported below.

\paragraph{Hypotheses and estimand.}
We state the directional hypothesis that future-conditioned training
improves episode success relative to current/no-future-head at fixed compute.
For paired episodes \(i=1,\ldots,N\), the primary estimand is
\(\Delta= N^{-1}\sum_i (Y_i^{\mathrm{future}}-Y_i^{\mathrm{current}})\),
where \(Y_i\in\{0,1\}\) is the terminal success indicator under shared initial
state and rollout noise.  A positive point estimate is insufficient on its own:
we require a confidence interval and a paired test.  Both requirements are
met by the primary comparison under the original all-300-step protocol:
CurrentJEPA-only succeeds on 730 pairs, FutureJEPA-only succeeds on 733,
the estimate is $+0.03$ percentage points, the exact McNemar test gives
$p=0.958$, and the bootstrap 95\% interval is approximately
$[-0.73,+0.77]$ points.  We interpret this as no detectable aggregate
future-head gain in this single training run, not as proof that
the true effect is exactly zero.

\paragraph{Joint-embedding predictive architectures.}
JEPA-style methods predict representations rather than pixels, which can suppress nuisance detail while retaining task-relevant structure \cite{lecun2022path,assran2023jepa}.  V-JEPA and subsequent video representation work motivate using a frozen visual encoder as a reusable state abstraction \cite{bardes2024vjepa}.  Our setting differs in that the prediction target is consumed by a VLA policy and is trained jointly with a flow-matching action objective.  It
complements world-model approaches that learn latent dynamics for planning
and control \cite{ha2018worldmodels,hafner2023dreamer}: our predictor is
evaluated as supervision for a frozen-encoder policy, not as a planner.

JEPA-WAM couples a V-JEPA-space transition predictor to continuous action generation through a shared predictor \cite{lin2026jepawam}.  We use that line of work as architectural motivation, but report a VLM-conditioned DiT flow policy with an explicit current/no-future-head control; the two settings should not be treated as identical implementations or benchmarks.

The closest concurrent work is VLAFlow \cite{xia2026vlaflow}, a unified flow-matching framework for controlled comparison of VLA pre-training paradigms.  Its MindWPI paradigm is directly related to ours: a frozen V-JEPA~2 encoder \cite{bardes2025vjepa2} supplies current and future frame latents at a fixed offset, and the action expert is trained to predict the future latent with an MSE objective alongside the action loss.  On a heterogeneous 5{,}000-hour pre-training corpus, future latent alignment improves transfer, and its combination with language co-training (MindLWPI) yields the most stable results on LIBERO, LIBERO-Plus, and SimplerEnv
\cite{li2024simplerenv}---independent evidence that future-latent supervision can benefit VLA control.  Our study differs in aim and integration.  VLAFlow compares pre-training paradigms under a shared KV-cache DiT expert and a unified 14-dimensional action space, with evaluation after downstream fine-tuning; we isolate a minimal intervention at the policy stage on a single data distribution, with both the VLM and the V-JEPA encoder frozen.  Architecturally, their action expert itself produces the future prediction, whereas FutureJEPA attaches two independent residual heads (offsets 4 and 16, tokenwise MSE loss) to the shared trunk while preserving the native flow prefix and masking future queries from noisy action tokens, leaving the inference path unchanged.  We use episode-paired comparisons between policies with and without future supervision, rather than comparisons across pre-training data mixtures.

\paragraph{Vision-language-action policies.}
OpenVLA, \(\pi_0\), and related systems combine visual and language features with continuous, diffusion, or flow-based action decoders \cite{brohan2023rt2,kim2024openvla,octo2024,black2024pi0,chi2023diffusion}.  We retain the native flow action path to isolate the effect of future latent supervision.

\paragraph{Representation alignment and physical grounding.}
Adapter-based transfer and representation alignment are attractive when the visual backbone is frozen
\cite{nair2022r3m,majumdar2023vc1,karamcheti2023voltron}.  Our future objective is complementary to physical grounding and to PhyLatent-style invariance or counterfactual losses \cite{phylatent2026,psgjepa2026}.  Combining these objectives with future prediction remains a direction for future work.

\paragraph{Contributions.}
Our contributions are:
\begin{enumerate}
\item We introduce training-only residual heads for future visual
representations while retaining a VLM-conditioned DiT flow policy and its
inference-time action sampler.
\item We evaluate the incremental effect of future supervision with
episode-paired current-feature and no-JEPA controls across the complete
LIBERO-Plus benchmark. The future-head contrast is $-0.46$ percentage
points under the final mixed-cap protocol and $+0.03$ points under the
original all-300-step protocol, with no detectable aggregate benefit.
\item We study two forms of test-time adaptation: persistent and transient
context inferred from executed transitions, and updates to selected model
parameters. Paired target controls distinguish lower prediction loss from
better closed-loop control.
\end{enumerate}

\section{Method}
\subsection{Shared action path}
\label{sec:shared-action-path}
Let \(z_t\in\mathbb{R}^{S\times D_J}\) denote the current JEPA token map and let \(c_t\) denote the policy conditioning, including VLM features and robot state. JEPA tokens are normalized, pooled from \(S_{in}=64\) to \(S=32\), projected to the flow token width, and appended to the pretrained learned prefix. This preserves the original prefix tokens. Future queries are masked from noisy action tokens.

The VLM side of the conditioning enters through a separate path from the
JEPA tokens. The frozen vision-language model is run once per replan; its
final-layer output hidden states are passed through a trainable LayerNorm
and a trainable VLM self-attention block, and the result is supplied to the
action expert as cross-attention context. Every DiT block~\cite{peebles2023dit} in the
action expert then reads this single shared VLM representation through its
own cross-attention sublayer (each block forms its own key and value
projections), rather than receiving a per-layer bridged cache.  The JEPA
tokens, by contrast, enter inside the DiT self-attention sequence as the
appended prefix block described above.  The two streams therefore
reach the trunk through distinct attention mechanisms: VLM features via
per-block cross-attention, JEPA tokens via in-sequence self-attention.

For an action sequence \(a_{t:t+H-1}\), flow matching samples \(\epsilon\sim\mathcal{N}(0,I)\) and \(\tau\sim U[0,1]\), forms
\[
 x_\tau=(1-\tau)\epsilon+\tau a,\qquad v^*=a-\epsilon,
\]
and minimizes the native velocity loss \(\Flow=\|v_\theta(x_\tau,c_t)-v^*\|_2^2\) \cite{lipman2023flowmatching,liu2023rectifiedflow}, with the implementation's action mask and dimensional padding.

\subsection{Future heads}
The shared trunk emits a latent query state \(q_t\).  For each horizon \(h\), an independent linear head predicts a residual:
\[
 \Delta_h = g_h(\mathrm{LN}(q_t)),\qquad \hat z_{t+h}=z_t+\Delta_h.
\]
The \foursixteen{} variant uses \(h=4\) and \(h=16\).  The auxiliary loss is
\[
 \begin{aligned}
 \Lfuture &= \tfrac{1}{2}\ell(\hat z_{t+4},z_{t+4})
             + \tfrac{1}{2}\ell(\hat z_{t+16},z_{t+16}),\\
 \Ltotal &= \Flow + 0.1\Lfuture .
 \end{aligned}
\]
Here \(\ell\) is tokenwise mean-squared error. Missing future targets are
masked at the sample level. Target features are produced by the frozen
JEPA encoder and detached from the gradient graph, so the auxiliary loss
updates the trainable projection, shared trunk, and prediction heads while
keeping the target representation fixed.

\paragraph{Prefix routing.}
All reported policies retain the native learned prefix. For the
JEPA-conditioned policies, the DiT self-attention sequence is
\[
 x_t = [\,c_t^{\mathrm{state}};\; p_{\mathrm{native}};\; P(\mathrm{LN}(z_t))+e_{pos}\;;\; x_\tau\,],
\]
The projected JEPA tokens are placed after the native prefix and before
the noisy action tokens. The native prefix contains 32 learned tokens;
the JEPA map is pooled from 64 to 32 tokens before projection. Prefix and
future-latent queries cannot attend to noisy action tokens, while action
queries can read the complete prefix. This mask allows the action expert
to use the visual representation without exposing the future-prediction
branch to the action-noise input.

\subsection{Control definition}
CurrentJEPA uses the same conditioning sequence, flow loss, optimizer, and
inference path as FutureJEPA, with \(\Lfuture=0\). BaseFlow retains the
learned prefix but omits JEPA conditioning. Thus CurrentJEPA versus
FutureJEPA isolates the additional future-prediction objective, whereas
BaseFlow versus FutureJEPA changes both conditioning and supervision.

\subsection{Inference}
Future predictions are training-only in the default policy.  At inference the
auxiliary branch is absent and the action sampler runs the native four-step
Euler flow solver, identical across all three arms.

\paragraph{Euler solver.}
Each action chunk begins from Gaussian noise \(x_0\sim\mathcal{N}(0,I)\) in
the native action space \(\mathbb{R}^{H\times D_a}\) (\(H=16\), \(D_a=32\)).
The learned velocity field is integrated with four uniform steps of width
\(\Delta\tau=\tfrac{1}{4}\):
\[
 x_{\tau+\frac14} = x_\tau + \tfrac{1}{4}\,v_\theta(x_\tau, c_t, \tau),
 \qquad \tau\in\{0,\tfrac14,\tfrac12,\tfrac34\},
\]
so one chunk costs exactly four DiT forward passes.  The result is cropped
from the native width back to the environment width (\(32\to 7\) DoF),
de-normalized with the dataset statistics, and the gripper channel is
binarized by sign before execution.  The full 16-step chunk is then executed
open-loop for one replan interval (16 environment steps) before the next
chunk is sampled, under the episode cap of the selected evaluation protocol
(300 steps in the original all-300 protocol; 300/600 steps in the final
mixed-cap headline protocol).
Paired rollouts use matched initial action-noise streams.

\paragraph{Inference cost.}
All three policies use four Euler steps and the same replan interval.
The future heads are omitted at inference, so the auxiliary objective does
not add action-sampling forward passes. Measured wall-clock latency was
not independently benchmarked.

\section{Test-Time Adaptation}
\label{sec:fixed-state-retry}
\label{sec:fixed-noise-retry}
This section describes two forms of test-time adaptation (TTA)
\cite{sun2020ttt,wang2021tent,wang2022cotta}. Context adaptation optimizes a
small latent state while all network weights remain frozen; parameter
adaptation updates a selected module from recent transitions. Their online
update rules are described here, and offline training is collected in
Section~\ref{sec:data-training}.

The context-adaptation and bounded-retry route is named
\textbf{SaiVLA-JEPA-TCA} (test-time context adaptation). The separate
selected-parameter update route is named
\textbf{SaiVLA-JEPA-TPA} (test-time parameter adaptation). They are
alternative update scopes around the same base policy.

\subsection{Context-conditioned policy}

\begin{figure}[t]
\centering
\includegraphics[width=\linewidth]{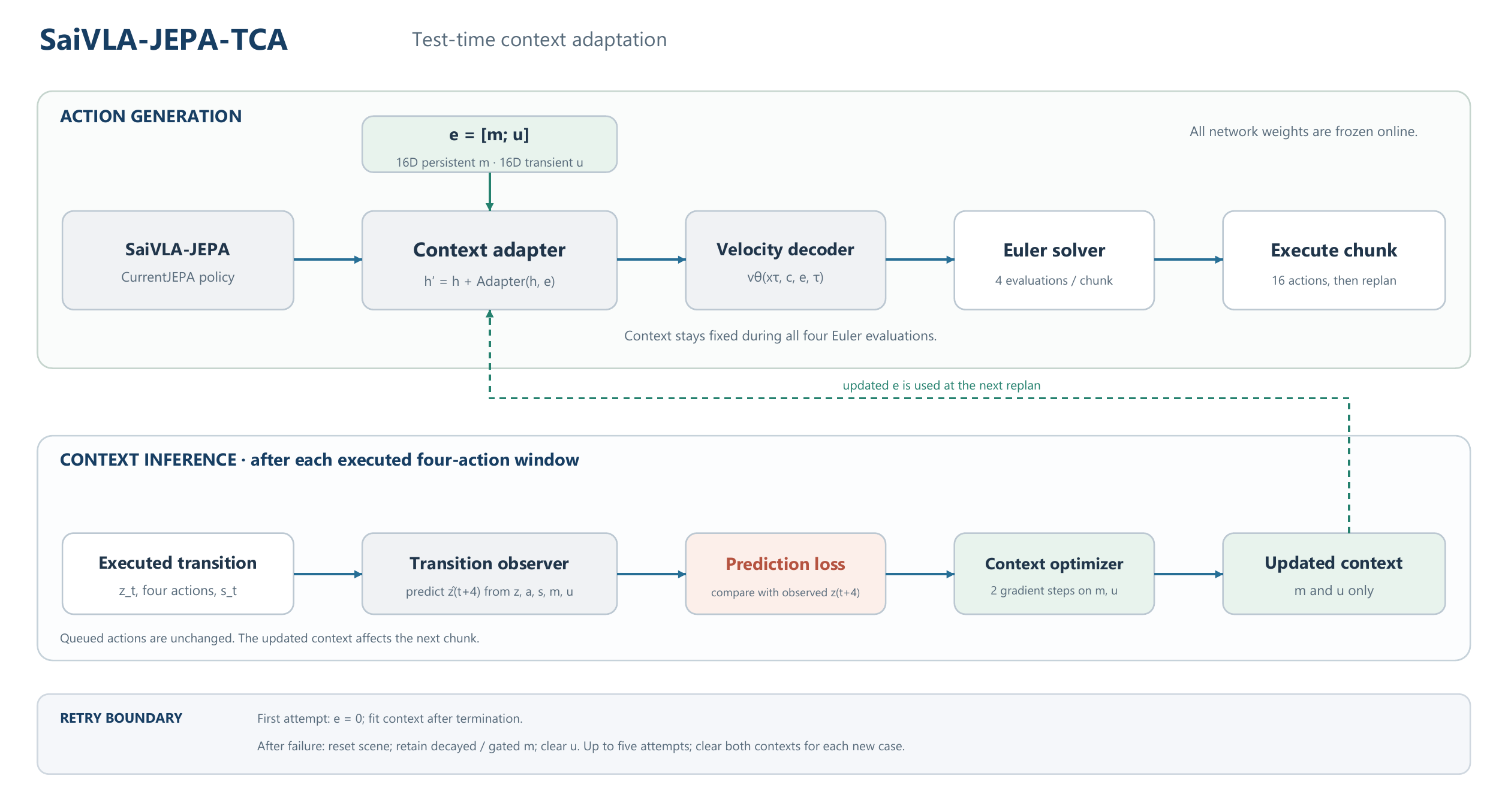}
\caption{SaiVLA-JEPA-TCA context pathway. The frozen SaiVLA-JEPA policy applies
the context adapter at each of four Euler evaluations using fixed
$e=[m;u]$. After every executed four-action window, a frozen observer
computes a latent prediction loss and two gradient steps update only $m$
and $u$ for the next action chunk. The initial attempt uses zero context
and defers fitting until it ends. On failure, $m$ is retained with decay
or evidence gating while $u$ is cleared. Network weights remain frozen
online.}
\label{fig:retry-tta-arch}
\end{figure}

\paragraph{Base policy.}
The complete retry evaluation uses CurrentJEPA trained for 25,000 steps
without future auxiliary supervision. A frozen vision-language model
supplies cross-attention features to a DiT action expert trained by flow
matching. The visual encoder is the ViT-g variant of V-JEPA~2.1
\cite{mur2026vjepa21}, with $384\times384$-pixel image inputs and
1,408-dimensional token features. At the DiT input, 32 learned prefix tokens
are retained and 32 projected JEPA tokens are appended
(Section~\ref{sec:shared-action-path}). The policy generates 16-action
chunks using four Euler steps.

\paragraph{Persistent and transient context.}
We separate adaptation into two 16-dimensional latent variables.
The persistent context $m$ is intended to retain information about
environment properties that remain useful across local transitions and
repeated attempts from the same initial state. The transient context $u$
is intended to capture short-lived discrepancies between the predicted
transition and the observation produced by the executed actions. It is
re-estimated for each recent window and is not transferred to the next
retry. This division separates an intended memory of the environment
from rapid local correction; it does not assume that either variable
has learned an interpretable physical parameter.
The policy receives their concatenation $e=[m;u]\in\mathbb{R}^{32}$.

\paragraph{Transition observer.}
A frozen observer predicts the visual feature reached after four executed
actions. Its input contains the current pooled JEPA feature
$z_t\in\mathbb{R}^{1408}$, four 7-dimensional actions $a_{t:t+3}$,
the robot state $s_t\in\mathbb{R}^{8}$, and the two contexts:
\begin{align*}
 x_t &= [z_t,\operatorname{vec}(a_{t:t+3}),s_t,m,u],\\
 \hat z_{t+4} &= z_t+N(\operatorname{LN}(x_t))+W_m m+W_u u.
\end{align*}
The input width is 1,476. The network $N$ is a two-hidden-layer MLP with
widths $1476\to512\to512\to1408$, SiLU activations, and an intermediate
LayerNorm; $W_m$ and $W_u$ are bias-free context projections.
After execution, the frozen visual encoder produces the observed target
$z_{t+4}$. Minimizing
$L=\|\hat z_{t+4}-z_{t+4}\|_2^2$ with respect to $m$ and $u$ adapts the
context to the transition that actually occurred. Both networks remain
frozen online.

\begin{samepage}
\paragraph{Action adapter at each Euler step.}
For one 16-action chunk, the flow sampler starts from a noisy action tensor
$x_0\in\mathbb{R}^{16\times32}$ and evaluates the velocity field at four
times $\tau\in\{0,1/4,1/2,3/4\}$. At each evaluation, the DiT receives the
current noisy action $x_\tau$ and time $\tau$, together with the VLM and JEPA
conditioning and robot state, and returns a 1,024-dimensional hidden state
for each action token. The adapter receives each such state
$h_\tau\in\mathbb{R}^{1024}$ together with $e=[m;u]\in\mathbb{R}^{32}$,
and returns the adapted hidden state $h'_\tau$ before the velocity decoder:
\begin{align*}
 q_\tau &= \operatorname{GELU}
 (W_{\rm down}\operatorname{LN}(h_\tau)+b_{\rm down}),\\
 g &= \tanh(W_e e),\\
 h'_\tau &= h_\tau+W_{\rm up}(q_\tau\odot g).
\end{align*}
Across the 16 action tokens, the hidden states form
$H_\tau\in\mathbb{R}^{16\times1024}$. The same 32-dimensional $e$ is
broadcast across these tokens, and the adapter returns
$H'_\tau\in\mathbb{R}^{16\times1024}$; the velocity decoder then produces
$v_\theta\in\mathbb{R}^{16\times32}$.
\end{samepage}
The bottleneck width is 128, with 268,416 parameters trained offline.
Since $W_e$ is bias-free, $e=0$ gives $h'_\tau=h_\tau$.
The velocity decoder maps $h'_\tau$ to $v_\theta(x_\tau,c_t,e,\tau)$, and the
sampler advances the action by $x_{\tau+1/4}=x_\tau+\tfrac14v_\theta$;
the resulting action chunk is cropped to the seven executed dimensions.
The same context $e$ is supplied at all four evaluations of this chunk:
the action tensor and timestep change, but $m$ and $u$ remain fixed. The
adapter is therefore applied four times, while no context-gradient step
occurs inside the sampler. Context updates are made only after an executed
four-action window and affect the next action chunk.
The $\tanh$ gate bounds its modulation coefficients; the complete retry
evaluation does not impose a hard bound on the final action residual.

\subsection{Online context updates and retries}
\label{sec:context-update-timing}

\paragraph{Within an attempt.}
The policy replans after executing a 16-action chunk. The observer divides
that execution into four-action windows and, once each window has completed,
performs two gradient steps on $m$ and $u$. Actions remaining in the queued
chunk are not recomputed. The updated context is read when the next
16-action chunk is generated, and is used in all four Euler evaluations
of that new plan.
The persistent context starts from its previous value and accumulates
updates within the attempt; the transient context is reset to zero before
fitting each window. With $\eta_m=\eta_u=100$, the updates are
\[
 m\leftarrow m-0.1\eta_m\nabla_m L,\qquad
 u\leftarrow u-\eta_u\nabla_u L.
\]
Each variable is projected separately to an $\ell_2$ ball of radius 3.
Context inference follows the executed transition windows while all network
weights remain frozen. The update therefore operates on a transition
window, rather than on a complete rollout or an Euler denoising step.

\paragraph{Initial attempt.}
For the first attempt, all methods act with $e=0$ throughout the rollout.
The observer windows are collected during execution and their context
updates are deferred until the attempt ends. Consequently, the inferred
terminal context can initialize a retry but cannot affect the first
attempt's actions. This is an exception to the within-attempt update
schedule used from the second attempt onward.

\paragraph{Between attempts.}
A failed attempt restores the same simulator initial state and reuses
the same flow-noise stream. The transient context is cleared.
With fixed decay, the next attempt retains a scaled version of the final persistent
context:
\[
 m_{r+1}^{\rm start}=\operatorname{clip}_3(0.1m_r^{\rm end}).
\]
With evidence gating, the update separates the initial memory from the
within-attempt innovation $\Delta m_r=m_r^{\rm end}-m_r^{\rm start}$:
\[
 m_{r+1}^{\rm start}
 =\operatorname{clip}_3(0.1m_r^{\rm start}+\alpha_r\Delta m_r).
\]
Here $\operatorname{clip}_3$ projects to radius 3, and $\alpha_r$ depends
on transition-prediction improvement. For each window, the evidence score is
\[
 g=\operatorname{clip}_{[-1,1]}\!\left(
 \frac{L_{\rm old}-L_{\rm new}}{\max(|L_{\rm old}|,10^{-12})}\right).
\]
The two losses compare the attempt-start memory with the accumulated memory
before fitting that window, using the same transient context. The first
window is excluded. The innovation is rejected unless at least four
evidence windows exist, $\|m_{\rm end}\|\le2.7$, and the cosine similarity
between $m_{\rm start}$ and $\Delta m$ is at least $-0.25$ whenever both
vectors are nonzero.
Eligible evidence selects $\alpha$ as 0.1, 0.25, or 0.5 for increasing
thresholds on mean gain and the fraction of positive-gain windows:
$(0.015,0.60)$, $(0.04,0.70)$, and $(0.08,0.80)$, respectively.
If no threshold is met, $\alpha=0$ and the previous memory still decays.
Task success is used only to stop retries, not to fit the context.
Each case allows at most five attempts and stops at first success.
All context is reset between cases; no task-to-task transfer is evaluated.

\begin{table}[t]
\begin{reporttable}[0.88\linewidth]
\caption{Settings for the complete retry evaluation.}
\label{tab:retry-full-contract}
\begin{tabular}{@{}>{\raggedright\arraybackslash}p{0.25\linewidth}>{\raggedright\arraybackslash}p{\dimexpr0.75\linewidth-2\tabcolsep\relax}@{}}
\toprule
Component & Setting \\
\midrule
Policy & CurrentJEPA after 25,000 training steps \\
Observer and adapter & from the same training partition; fixed at evaluation \\
Initial conditions & same simulator state and flow-noise stream across methods \\
Attempt budget & baseline: 1; context methods: at most 5, stopping at success \\
Episode caps & Spatial/Object: 300; Goal/LIBERO-10: 600 per attempt \\
Action generation & 16-action chunks; 4 Euler evaluations per chunk \\
Context update & every completed 4-action window; 2 steps on $m,u$ \\
Initial attempt & zero context; context fitting deferred until termination \\
Reset & $u$ cleared between attempts; both $m,u$ cleared between cases \\
\bottomrule
\end{tabular}
\reporttablenote{The baseline and retry methods have different maximum attempt budgets.}
\end{reporttable}
\end{table}

\subsection{Parameter adaptation}
\label{sec:param-tta}
We call this separate route \textbf{SaiVLA-JEPA-TPA}. It updates a small parameter subset rather than the latent
contexts. In the standard setting, at each 16-step replan boundary, the executed chunk and new
observation form a transition. A frozen target encoder supplies the latent
target, and a single AdamW step with learning rate $10^{-5}$ and gradient
clip 0.1 produces a candidate update. Acceptance requires a target-quality
comparison against shuffled targets and action-change checks, including a
chunk-drift bound of 0.005. A rejected update is rolled back. Parameters
are restored at the start of each episode; a separate persistent-replay
experiment evaluates whether an update from one attempt remains useful
when replayed without further adaptation. The interval study instead
accumulates transitions over longer windows before updating.

We use the following controls:
\begin{itemize}
\item \textbf{No update:} execute the frozen policy.
\item \textbf{Matched target:} update using the observed future representation
of the executed transition.
\item \textbf{Shuffled target:} replace that target with one from a matched
donor episode, holding the observation, state, executed actions, and action
noise fixed.
\item \textbf{Zero input:} remove JEPA conditioning. In the joint
meta-adaptation study, the corresponding control instead uses a zero-valued
future target; the table note identifies that case.
\end{itemize}
These controls distinguish useful adaptation from generic perturbation.
Since no expert action labels are available online, a lower transition loss
or an accepted update does not by itself establish a control benefit.

\section{Data and Training}
\label{sec:data-training}
\subsection{Data and target construction}
The policy is trained on 14,347 demonstration episodes. Each training
window contains the current visual observation, robot state, a 16-action
chunk, and future JEPA targets at offsets 4 and 16 when available.
Missing future targets are masked rather than replaced by zeros.

Evaluation covers 10,030 LIBERO-Plus episodes across four suites
\cite{liu2023libero,fei2025liberoplus}: 2,402 Spatial, 2,518 Object,
2,591 Goal, and 2,519 LIBERO-10 episodes. The final success-rate tables use
300-step caps on Spatial/Object and 600-step caps on Goal/LIBERO-10.
The primary paired tests and completion-time analysis use the original
300-step cap on all suites. Each comparison uses one trial per task,
matched initial states, and matched flow-noise streams. Fixed random seeds
are used for training, data partitioning, and paired evaluation.

\subsection{Policy training}
All three policies use an effective batch size of 64 and 25,000 optimizer
steps. The learning rate warms up for 500 steps to $10^{-4}$ and then
follows cosine decay \cite{loshchilov2017sgdr} over the remaining training
budget. Thus the cosine schedule describes how the learning rate changes
during optimization; it is separate from the four Euler steps used to
sample each action chunk at inference. The VLM and visual encoder remain
frozen while the flow action expert and JEPA projection are trained.
CurrentJEPA and BaseFlow use the flow loss only. FutureJEPA adds the
residual future-latent MSE at offsets 4 and 16, with total weight 0.1 split
equally between the two heads.

At each optimization step, the model encodes the current observation and
available future frames, pools the current JEPA map from 64 to 32 tokens,
and projects these tokens into the action expert. It samples a flow time
and Gaussian action noise, computes the velocity target, and backpropagates
the action and available future losses. The future-target encoder is
detached from the gradient graph. At deployment, only the action branch
is evaluated.

\subsection{Training the transition observer}
The observer and context adapter are trained separately from the policy.
We train two observer--adapter pairs using independently randomized
training/validation partitions. Each observer partition contains 24
training and six validation episodes per task, with four windows per
episode. These are data partitions, not additional benchmark evaluation
runs. The complete retry evaluation uses the second trained pair
throughout; the earlier diagnostics report results from both pairs
where applicable.

Support windows from different episodes of the same task infer persistent
context, encouraging $m$ to explain information shared across episodes.
A separate non-query window from the query episode infers transient
context $u$, allowing local correction before predicting a held-out
four-action query. If $m_A$ and $m_B$ are persistent contexts inferred
from two support examples, the query uses their mean and a consistency
loss penalizes $\operatorname{mean}((m_A-m_B)^2)$.
The objective is
\begin{align*}
 \mathcal L_{\rm obs}={}&L_{\rm query}+0.2L_{\rm support}
 +0.2L_{\rm consistency}+2L_{\rm contrast}\\
 &+10^{-6}\operatorname{mean}(m^2)
 +10^{-5}\operatorname{mean}(u^2).
\end{align*}
The support term sums two persistent-support losses and one transient
support loss. The contrastive term averages hinge losses against
mismatched-memory and zero-memory controls, with margin $10^{-4}$.

Observer training uses 5,000 AdamW steps \cite{loshchilov2019adamw},
batch size 128, learning rate $3\times10^{-4}$, weight decay $10^{-5}$,
$\beta=(0.9,0.99)$, and gradient clipping at 1. Offline context inference
uses two inner steps with learning rate 300 for each context.
Validation is performed every 250 steps, and model selection maximizes
the smaller of the gains over the zero-memory and mismatched-memory
controls.

\subsection{Training the context adapter}
The adapter is trained with a frozen FutureJEPA policy after 3,000
training steps. The same adapter is subsequently used with the
25,000-step CurrentJEPA policy in the complete retry evaluation.
This transfer between policy checkpoints is retained as a limitation;
the adapter is not retrained on the evaluation episodes.

Only adapter weights are optimized; the policy and observer remain
frozen. The loss combines flow-velocity MSE under the inferred context
with margin-$10^{-3}$ hinge terms against zero context, shuffled context,
random context, and context without the transient component.
Training uses 320 training and 80 validation samples, batch size 8,
5,000 AdamW steps, learning rate $5\times10^{-4}$, and weight decay
$10^{-5}$. The selected adapter maximizes the minimum validation gain
over zero, shuffled, and random context. The complete evaluation uses
the observer and adapter from the same training partition.

\subsection{Evaluation statistics}
For the primary paired comparison under the all-300-step protocol, we
report success counts, candidate-minus-control percentage points, a
bootstrap 95\% interval \cite{efron1993bootstrap}, and exact McNemar's
test \cite{mcnemar1947}. The final mixed-cap success rates are descriptive
and do not inherit that protocol's interval or $p$-value. Separate
ablations specify their own evaluation settings. Latent prediction is
compared with a repeat-current predictor; a lower latent loss is not,
by itself, evidence of better closed-loop control.

\section{Experiments and Evidence}
\label{sec:experiments}
\subsection{Main paired comparison}
The three policies are evaluated on the same 10,030 LIBERO-Plus episodes.
BaseFlow omits JEPA conditioning, CurrentJEPA uses current JEPA features
without future supervision, and FutureJEPA additionally predicts future
representations. The policies share their training schedule, data windows,
and effective batch size; paired episodes use matched initial states and
action noise.

Under the original protocol, with a 300-step limit for every suite, FutureJEPA succeeds on
7,924 of 10,030 episodes (79.00\%), compared with 6,628 (66.08\%) for
BaseFlow. This paired difference is +12.92 percentage points (95\% bootstrap
interval: +12.04 to +13.82; exact McNemar $p\approx2.5\times10^{-174}$),
but it combines JEPA conditioning with future supervision.

Compared with CurrentJEPA (7,921 successes, 78.97\%), FutureJEPA records
three more successes, a difference of +0.03 points ($p=0.958$; 95\% interval:
$[-0.73,+0.77]$). We find no detectable aggregate benefit in success from
future supervision in this training run.

Table~\ref{tab:fullplus} reports the final mixed-cap rates: 71.63\% for
BaseFlow, 82.45\% for CurrentJEPA, and 81.99\% for FutureJEPA.  The
paired statistics above use the original all-300-step protocol; mixed-cap
contrasts are descriptive.

\begin{table}[t]
\begin{reporttable}
\caption{LIBERO-Plus success counts (rates): final protocol-matched evaluation.}
\label{tab:fullplus}
\begin{tabular*}{\linewidth}{@{\extracolsep{\fill}}lccc@{}}
\toprule
Suite (episodes, cap) & BaseFlow & CurrentJEPA & FutureJEPA \\
\midrule
Spatial (2{,}402, 300) & 1958 (81.52\%) & 2072 (86.26\%) & \textbf{2097 (87.30\%)} \\
Object (2{,}518, 300) & 2072 (82.29\%) & \textbf{2190 (86.97\%)} & 2177 (86.46\%) \\
Goal (2{,}591, 600) & 1816 (70.09\%) & 2057 (79.39\%) & \textbf{2090 (80.66\%)} \\
LIBERO-10 (2{,}519, 600) & 1338 (53.12\%) & \textbf{1951 (77.45\%)} & 1860 (73.84\%) \\
\midrule
Overall (10{,}030, mixed) & 7184 (71.63\%) & \textbf{8270 (82.45\%)} & 8224 (81.99\%) \\
\bottomrule
\end{tabular*}
\reporttablenote{The final protocol uses 300-step caps on LIBERO-Spatial/Object and 600-step caps on LIBERO-Goal/LIBERO-10, with one trial per task and a 16-action replanning interval. Episode keys, initial states, and action-noise streams are matched across policies. Paired all-300-step statistics are reported in the text. Bold marks the highest rate in each suite; ties are bolded.}
\end{reporttable}
\end{table}

\subsection{Completion-time efficiency}
Success rate is not the only axis on which the arms differ.  Under the
original all-300-step protocol, we additionally
compare how quickly each policy completes the episodes it succeeds on, using
the executed-step counts in the episode records.  Raw per-suite means are
confounded because each arm succeeds on a different episode set, so the
primary statistic is paired: for each pair of arms we restrict to episodes
both arms complete and compare step counts episode by episode.

On the 7{,}191 episodes completed by both JEPA-conditioned arms,
CurrentJEPA finishes earlier than FutureJEPA by 4.34 steps on average
(median 3 steps; Wilcoxon signed-rank \cite{wilcoxon1945} \(p\approx1.2\times10^{-118}\);
bootstrap 95\% interval \([-4.95, -3.74]\) steps for the
CurrentJEPA-minus-FutureJEPA difference, 10{,}000 resamples;
CurrentJEPA is faster on 65.7\% of paired episodes).  The advantage is
consistent across all four suites and grows with suite horizon: \(-0.58\)
steps on LIBERO-Spatial, \(-5.61\) on LIBERO-Object, \(-5.32\) on
LIBERO-Goal, and \(-6.55\) on LIBERO-10 (negative means CurrentJEPA
faster).  Against BaseFlow, CurrentJEPA has a paired mean advantage of 7.01
steps, with BaseFlow slower on 56.6\% of jointly successful episodes.  The
available analysis does not retain a signed-rank result for this
contrast, so we do not report an exact $p$-value.  The BaseFlow-versus-FutureJEPA
contrast is the only unstable one: the paired mean favors FutureJEPA by
3.19 steps (bootstrap 95\% interval \([+2.40, +3.98]\)), but the
signed-rank test is not significant (\(p\approx0.40\)) because a majority
of paired episodes (55.5\%) in fact favor BaseFlow---the mean is carried
by a minority of episodes where BaseFlow stalls near the step cap before
recovering.

This indicates a modest efficiency benefit from current JEPA
conditioning: both JEPA-conditioned arms tend to finish sooner than
BaseFlow, CurrentJEPA is the fastest completer of the three, and the
future-latent objective adds a small completion-time cost (about four
steps against a 300-step cap) rather than a gain at this budget.  Step
counts are recorded only up to success, so this analysis says nothing
about failure episodes, which all terminate at the 300-step cap under
this protocol.

\subsection{Zero-shot JEPA conditioning and projection transfer}
\label{sec:zeroshot-full}
The three-arm comparison associates the large observed BaseFlow gap with the
presence of JEPA conditioning, but does not identify where that gain enters
the network.  We therefore isolate the two integration sites with a zero-shot
construction: the BaseFlow 25,000-step checkpoint is executed under the CurrentJEPA
configuration, so JEPA tokens are fed to a trunk that was never trained with
them, and only the JEPA projection $P$ differs between arms. We compare
the untrained random projection retained from BaseFlow, the trained
projection transferred from CurrentJEPA, and a zero-padded identity
mapping. No parameter is updated during evaluation. Each arm covers
10{,}030 episodes under the same per-suite limits as the trained reference
policies: 300 steps on Spatial/Object and 600 on Goal/LIBERO-10, with one
trial per task and replanning every 16 actions. All comparisons are paired
by episode. This experiment measures how projection transfer changes
performance when JEPA features are introduced to an untrained input path;
it does not isolate the effect of future supervision.

\begin{table}[t]
\begin{reporttable}
\caption{Full-scale zero-shot JEPA-injection comparison: success rates (\%).}
\label{tab:zeroshotfull}
\begin{tabular*}{\linewidth}{@{\extracolsep{\fill}}lrrrrr@{}}
\toprule
Arm & Spatial & Object & Goal & LIBERO-10 & Overall \\
\midrule
BaseFlow & 81.52 & 82.29 & 70.09 & 53.12 & 71.63 \\
Random projection & 82.72 & 81.25 & 69.32 & 61.18 & 73.48 \\
Trained projection & 83.68 & 82.33 & 74.91 & 64.39 & 76.23 \\
Identity projection & 80.60 & 81.37 & 67.04 & 55.38 & 70.96 \\
\midrule
CurrentJEPA & 86.26 & \textbf{86.97} & 79.39 & \textbf{77.45} & \textbf{82.45} \\
FutureJEPA & \textbf{87.30} & 86.46 & \textbf{80.66} & 73.84 & 81.99 \\
\bottomrule
\end{tabular*}
\reporttablenote{Each arm covers 10{,}030 paired episodes under the final
per-suite limits. The three projection variants use the BaseFlow trunk,
which was trained without JEPA tokens. CurrentJEPA and FutureJEPA retain
their own projections and trunks, trained without and with the future
objective, respectively. Bold marks the highest rate in each column among all evaluated arms.}
\end{reporttable}
\end{table}

Feeding JEPA tokens through the untrained random projection yields a
small but significant improvement over BaseFlow ($+1.85$ points; McNemar
discordant pairs 1116/930, $p=4.3\times10^{-5}$) that is concentrated almost
entirely on LIBERO-10 ($+8.06$ points) and absent or slightly negative on the
other three suites: the zero-shot token path is helpful but
suite-dependent. Transplanting the trained
projection adds $+2.75$ points (1141/865, $p=7.8\times10^{-10}$),
concentrated on Goal ($+5.60$) and LIBERO-10 ($+3.22$), and the identity
projection remains significantly worse than the random one ($-2.52$ points,
$p=1.3\times10^{-8}$), consistent with an untrained trunk interface favoring
a random basis over a deterministic passthrough in this protocol.  The
residual gap to CurrentJEPA, $+6.22$ points (1248/624,
$p=7.1\times10^{-48}$), is also consistent with a contribution from
trunk-side training.  Among the additional $+8.97$ points beyond the
random-injection arm, projection transplantation accounts descriptively for
31\% and the remaining trained-policy contrast for 69\% in this particular
zero-shot construction; these percentages are not estimates of distinct
causal contributions. These
protocol-specific contrasts are consistent with a larger gap after trained
policy integration than after transplanting the projection alone; they do
not isolate which parameter updates produce it.

\subsection{Training-side diagnostics}

\paragraph{Latent prediction.}
In a separate 400-target diagnostic at step 1500, the \(h=16\) prediction
MSE is 0.10295 versus 0.11177 for repeating the current representation.  At
\(h=4\), the values are 0.04152 and 0.04134.  This diagnostic uses an
adapter-only meta-adaptation model initialized from an earlier 3,000-step training run; it is not a direct validation of the
headline 25,000-step FutureJEPA checkpoint.  It shows that this diagnostic branch is
not a constant repeat operation at the long horizon, while the short horizon
is close to the identity baseline. Matched and shuffled conditioning have
nearly identical action L1 error, so this diagnostic does not establish
that the conditioning improves action prediction through its temporal content.

\paragraph{Action-loss parity after 25,000 training steps.}
The 25,000-step training logs allow the action-loss comparison to be made at the
full training budget for all three arms.  BaseFlow, CurrentJEPA, and
FutureJEPA each contain 1,001 records on the same step grid and the
same data windows, effective batch size 64, cosine schedule, and image-flip
setting.  We compare \(\Flow\) directly because the FutureJEPA total
\(\Ltotal=\Flow+0.1\,\Lfuture\) includes auxiliary latent supervision.

The three arms converge to nearly the same action loss: at the final logged
step the Flow losses are 0.03017 (BaseFlow), 0.03185 (CurrentJEPA), and
0.02873 (FutureJEPA), and the trailing 2,500-step means agree within 1.3\% of
one another.  This parity is expected: the Flow loss is an open-loop
surrogate scored on noisy interpolated training actions, where current-frame
features already explain most of the target variance.  Closed-loop rollout is
different in kind---execution is sequential, visited states are generated by
the policy itself, and small per-step differences compound---so loss parity
on the static surrogate is compatible with the large rollout gaps of
Table~\ref{tab:fullplus}.  The completed three-arm comparison associates the
observed gain with JEPA conditioning and finds no detectable aggregate
future-head benefit at this budget.

\paragraph{Weight-space similarity.}
The three final checkpoints are extremely close in weight space: flattening
the 323 shared trainable tensors of the flow head, pairwise cosine
similarity is 0.9686--0.9733, with CurrentJEPA and FutureJEPA the most
similar pair, and the bulk of the parameter mass (action encoder, action
decoder) barely moves apart.  The divergence concentrates where the recipes
differ: the JEPA projection drops to cosine 0.805--0.825 between FutureJEPA
and either other arm, and the future-latent head tensors in the BaseFlow and
CurrentJEPA checkpoints remain effectively at their random initialization
(cosine \(\approx 0\)), confirming that those parameters receive no gradient
without the future objective.  The full numeric similarity record is unavailable, so this is a provenance-limited diagnostic rather than
independently reproducible evidence.

\subsection{Effect of the future-loss weight}
\label{sec:lambda_sweep}
The FutureJEPA objective weights the auxiliary latent term at
\(\lambda=0.1\), and at the end of training the two components are
\(\Flow\approx0.035\) and \(\Lfuture\approx0.054\), so the future signal
contributes only about 13\% of the scalar total.  One possible account of the
small observed contribution of the future objective
(Table~\ref{tab:fullplus}) is therefore that the signal is present but
under-weighted.  To test this directly we trained two further full 25,000-step
\foursixteen{} policies under the same training settings (effective
batch 64, the same pretrained model, data, and schedule), raising the
latent weight to \(\lambda=0.3\) and \(\lambda=1.0\); the latent term then
accounts for roughly 26\% and 50\% of the total objective respectively.
Each arm was evaluated on all 10,030 LIBERO-Plus episodes using the
limits in Table~\ref{tab:fullplus}: 300 steps on Spatial/Object and 600
on Goal/LIBERO-10.

Table~\ref{tab:lambda_sweep} reports the rollout results.  The training-side
effect of the sweep is unambiguous: the latent heads improve monotonically
with \(\lambda\)---the validation \(h{=}16\) MSE drops by 18\% at \(\lambda=0.3\) and by
33\% at \(\lambda=1.0\) relative to \(\lambda=0.1\)---and the action branch is
not harmed, with final Flow losses of 0.0305 and 0.0247 for the two new arms
bracketing the original 0.0287.  The closed-loop effect, however, is small
and concentrated: the mixed-cap aggregate changes are +0.71 points for
\(\lambda=0.3\) (exact McNemar \(p{=}0.0552\)) and +0.81 points for
\(\lambda=1.0\) (\(p{=}0.0361\)), concentrated on LIBERO-Object
(\(p<10^{-3}\) at both weights), with no nominal movement on Spatial or
Goal and no change in steps-to-success (LIBERO-Goal 122--126 steps across
all three weights).  The two raised-weight arms do not differ detectably
from each other (\(p{=}0.810\)) or from CurrentJEPA overall, and both remain
far above BaseFlow on every suite.  All p-values here are exact two-sided
McNemar values recomputed from the archived discordant counts; they are
nominal and unadjusted for the multiple suite/weight comparisons.

\begin{table}[t]
\begin{reporttable}[0.88\linewidth]
\caption{Loss-weight ablation on LIBERO-Plus: success rates (\%).}
\label{tab:lambda_sweep}
\begin{tabular*}{\linewidth}{@{\extracolsep{\fill}}lcrrrrr@{}}
\toprule
Arm & $\lambda$ & Spatial & Object & Goal & LIBERO-10 & Overall \\
\midrule
BaseFlow    & --- & 81.52 & 82.29 & 70.09 & 53.12 & 71.63 \\
CurrentJEPA & --- & 86.26 & 86.97 & 79.39 & \textbf{77.45} & 82.45 \\
FutureJEPA & 0.1 & \textbf{87.30} & 86.46 & 80.66 & 73.84 & 81.99 \\
\midrule
FutureJEPA & 0.3 & 86.43 & 88.48 & 80.16 & 75.98 & 82.70 \\
FutureJEPA & 1.0 & 86.64 & \textbf{89.40} & \textbf{81.09} & 74.32 & \textbf{82.80} \\
\bottomrule
\end{tabular*}
\reporttablenote{The lower block contains the two newly trained arms;
the upper block contains the three reference policies. All arms use the
limits in Table~\ref{tab:fullplus}, one trial per task, a 16-action
replanning interval, and no image flip. Each new arm completed all
10{,}030 episodes. Bold marks the highest rate in each column.}
\end{reporttable}
\end{table}

The sweep does not support simple under-weighting as the sole explanation:
a three- to ten-fold increase in \(\lambda\) produces a monotonic
improvement in latent prediction quality but at most about one point of
overall rollout success.  This pattern is consistent with a downstream
representation-to-control bottleneck, but does not identify the action
decoder as the unique cause.

\subsection{Retry-time context adaptation}
\label{sec:retry-complete}
On the complete four-suite cohort of 10,030 cases, the frozen policy
reaches 8,124 successes in one attempt. With up to five attempts,
fixed-decay context reuse reaches 8,914 and evidence-gated context reuse
reaches 8,901. These are results for bounded retries with unequal
attempt budgets, rather than estimates of a pure memory effect.

Each case is one task variant and selected initial state. The three
methods start from the same simulator state and flow-noise stream.
The no-context baseline executes one rollout; both context methods stop at their
first success or after five attempts. The denominator remains 10,030 cases.

\begin{table}[t]
\begin{reporttable}
\caption{Complete four-suite retry evaluation.}
\label{tab:retry-complete}
\begin{tabular*}{\linewidth}{@{\extracolsep{\fill}}lrrccc@{}}
\toprule
Suite & Cases & Cap & No context & Fixed decay & Evidence gate \\
\midrule
Spatial & 2402 & 300 & 2050 (85.35\%) & \textbf{2185 (90.97\%)} & \textbf{2185 (90.97\%)} \\
Object & 2518 & 300 & 2171 (86.22\%) & 2257 (89.63\%) & \textbf{2263 (89.87\%)} \\
Goal & 2591 & 600 & 2075 (80.08\%) & 2272 (87.69\%) & \textbf{2289 (88.34\%)} \\
LIBERO-10 & 2519 & 600 & 1828 (72.57\%) & \textbf{2200 (87.34\%)} & 2164 (85.91\%) \\
\midrule
Overall & 10030 & mixed &
8124 (81.00\%) & \textbf{8914 (88.87\%)} & 8901 (88.74\%) \\
\bottomrule
\end{tabular*}
\reporttablenote{Cells give counts (success rates). The cap is per attempt.
The no-context baseline receives one attempt; both context methods receive
up to five. Bold marks the highest success rate per suite; ties are both shown.}
\end{reporttable}
\end{table}

Relative to the within-experiment baseline, fixed decay gains 790 successful cases
(+7.88 percentage points), and evidence gating gains 777 (+7.75 points).
The paired cumulative outcomes are 802 successes unique to fixed decay versus
12 unique to the baseline, and 792 unique to evidence gating versus 15 unique
to the baseline. In the equal-budget comparison between the two context methods,
8,759 cases succeed with both, 155 with fixed decay only,
142 with evidence gating only, and 974 with neither
(exact McNemar $p=0.4863$). The aggregate evidence does not establish
an advantage for the evidence-dependent memory rule.

\paragraph{Success and cost.}
With fixed decay, cumulative success counts at budgets 1--5 are
8,130, 8,636, 8,766, 8,861, and 8,914. With evidence gating, the counts are
8,110, 8,591, 8,744, 8,830, and 8,901.
Most additional recoveries occur on the second attempt.
The one-attempt baseline does not provide a five-attempt no-context curve.

\begin{table}[t]
\begin{reporttable}[0.82\linewidth]
\caption{Cumulative evaluation cost.}
\label{tab:retry-full-cost}
\begin{tabular*}{\linewidth}{@{\extracolsep{\fill}}lrrr@{}}
\toprule
Quantity & No context & Fixed decay & Evidence gate \\
\midrule
Total attempts & 10{,}030 & 15{,}757 & 15{,}875 \\
Mean attempts/case & 1.000 & 1.571 & 1.583 \\
Total environment steps & 2{,}165{,}688 & 4{,}705{,}637 & 4{,}779{,}987 \\
Mean steps/case & 215.92 & 469.16 & 476.57 \\
Context-update windows & 0 & 1{,}170{,}851 & 1{,}189{,}441 \\
\bottomrule
\end{tabular*}
\reporttablenote{Includes all attempts until success or the cap.
Environment steps do not include the cost of context optimization or offline training.}
\end{reporttable}
\end{table}

The two context methods use about 2.17 and 2.21 times the baseline's environment steps,
respectively.
A complete equal-budget no-context arm and a shuffled-context arm are
needed before the gain can be attributed specifically to inferred memory.

\begin{figure}[t]
\centering
\includegraphics[width=0.98\linewidth]{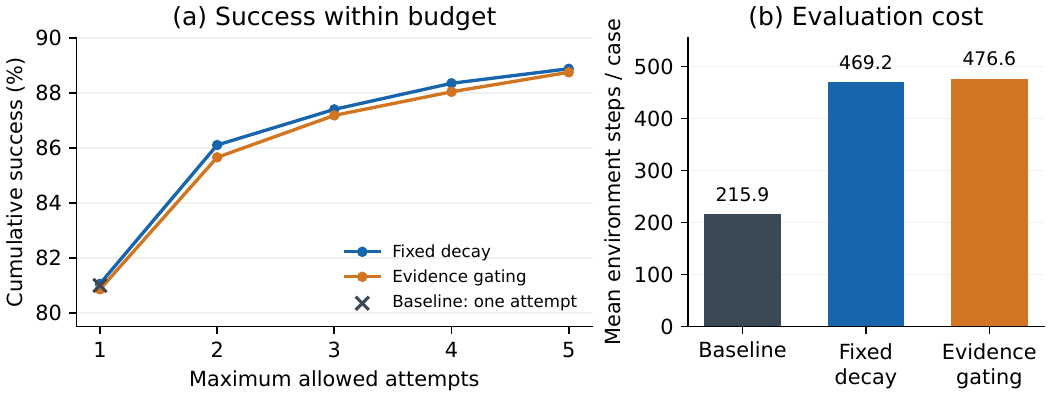}
\caption{Success and cost under bounded retries. Left: cumulative success
with up to five attempts, compared with the one-attempt baseline.
Right: mean cumulative environment steps per case.}
\label{fig:retry-full-budget}
\end{figure}

\paragraph{Pairing and evaluation variability.}
All 10,030 first attempts match in initial state, initial observation,
action noise, and first action. Nevertheless, full first-attempt
trajectories are not identical: action traces match the baseline in
8,709 cases with fixed decay and 8,704 with evidence gating. Since context remains zero during the
first attempt, its outcome differences are evaluation variability rather
than context effects. One later initial-observation mismatch remains;
excluding it gives 10,029 cases and 8,123, 8,913, and 8,900 successes for
the baseline, fixed decay, and evidence gating, respectively. The paired test
between the context methods is unchanged.
The retry baseline is 81.00\%, compared with CurrentJEPA's 82.45\% in the
main mixed-cap experiment; these are separate evaluation runs and their
baseline counts are not interchangeable.

\subsection{Additional retry comparisons}

\paragraph{Persistent context versus reset.}
An earlier FutureJEPA policy trained for 3,000 steps was evaluated on
16 selected failed context-conditioned records, corresponding to
14 unique initial states. Ten records came from the first observer
partition and six from the second. Carrying $m$ while resetting $u$
succeeds on 14/16 records; resetting both succeeds on 0/16, within
20 attempts of 300 steps. However, an initial zero-context rollout
already succeeds on 14/16, so the carry/reset comparison is not a 14/16
improvement over the baseline. With freshly sampled action noise,
both carry and reset reach 16/16. The cases therefore support an effect
of retained context, but do not isolate it from retry diversity.

\paragraph{Selected-case studies.}
A fixed schedule chosen on separate trials improves 54/80 baseline
successes to 69/80 with additional attempts. A separate 26-case
persistent-context pilot records 12 memory-only and one no-context-only
success over 338 attempts and 100,206 steps; its no-context stopping
budget depends on the memory method's trajectory. A 500-case Spatial
study with the 25,000-step CurrentJEPA policy records 396, 423, and 420
successes for the baseline, fixed decay, and evidence gating, respectively.
The two context methods have six versus three discordant successes
($p=0.5078$). These selected-case studies are not pooled with the complete
four-suite evaluation (Appendix~\ref{app:protocol-provenance}).

\paragraph{Partial varying-noise comparison.}
A separate comparison gives all methods up to five attempts and resamples
action noise. It covers 4,790 cases, with 4,329, 4,315, and 4,297
successes for the baseline, fixed decay, and evidence gating, respectively.
Four cases contain incomplete attempt
budgets; excluding them leaves 4,786 cases and 4,327/4,314/4,297 successes.
These partial results do not show an advantage for context reuse and
cannot substitute for a complete equal-budget evaluation.

\subsection{Parameter-level adaptation}
Table~\ref{tab:other_online} compares online updates against a no-update
policy and target controls. Task subsets and sample sizes differ across
rows, so the results are not pooled.

\begin{table}[t]
\begin{reporttable}[0.94\linewidth]
\caption{Success rates for parameter-level adaptation (\%).}
\label{tab:other_online}
\begin{tabular*}{\linewidth}{@{\extracolsep{\fill}}>{\raggedright\arraybackslash}p{0.32\linewidth}crrrr@{}}
\toprule
Method & Episodes $\times$ arms & \shortstack{No\\update} &
Matched & Shuffled & \shortstack{No JEPA /\\zero target} \\
\midrule
Visual-encoder LoRA & 20 $\times$ 4 & 25 & 30 & \textbf{35} & 30 \\
Shared DiT block & 40 $\times$ 3 & \textbf{32.5} & 27.5 & \textbf{32.5} & --- \\
Shared visual--action adapter & 20 $\times$ 3 & \textbf{30} & \textbf{30} & \textbf{30} & --- \\
Shadow updates\textsuperscript{a} & 80 $\times$ 2 & \textbf{65} & \textbf{65} & --- & --- \\
Joint visual--DiT adaptation & 80 $\times$ 4 & 72.5 & 72.5 & \textbf{73.75} & \textbf{73.75} \\
Persistent replay\textsuperscript{b} & 616 $\times$ 5 & 66.2 & 66.2/66.4 & \textbf{67.2/68.3} & --- \\
Multiple update intervals\textsuperscript{c} & 2400 $\times$ 4 & 64.2 & 64.2 & \textbf{65.3} & 64.3 \\
\bottomrule
\end{tabular*}
\reporttablenote{Matched and shuffled denote the temporal target used for
updates. The last column removes JEPA conditioning, except in joint
adaptation, where it uses a zero-valued future target. ---: condition not evaluated.
Initial states and action-noise streams are paired within each row. Bold marks the highest rate within each study; ties are bolded.
\textsuperscript{a}\,Updates are computed but candidate actions are not
executed; success therefore matches the no-update policy by construction.
\textsuperscript{b}\,After 500 offline optimization steps, each target condition
is evaluated on the initial rollout and again with its adapted parameters
retained, giving the two rates.
\textsuperscript{c}\,After the same offline initialization, pools 800 episodes
per arm at each of three update intervals (50, 100, and 150 steps).}
\end{reporttable}
\end{table}

\begin{samepage}
\subsubsection{Update location and outcome}
The experiments ask where a transition-prediction gradient can be applied
without degrading closed-loop control:
\begin{itemize}
\item \textbf{Visual encoder or projection.} Updates change visual
conditioning before it enters the action expert. Some prediction losses
improve, but success does not consistently exceed the no-update or target
controls.
\item \textbf{Shared DiT block or adapter.} Updates act more directly on
action hidden states. The 40-episode block update reaches 27.5\% versus
32.5\% without updates. Larger replay and interval studies also do not
separate matched-target adaptation from the controls.
\item \textbf{Joint and shadow updates.} Joint visual--DiT adaptation
matches the no-update policy on 58/80 episodes. Shadow evaluation computes
and checks updates while executing only frozen-policy actions; its equal
success rate tests neither the utility nor safety of the candidate actions.
\end{itemize}
These results distinguish a working latent-update mechanism from a useful
control adaptation rule. No tested parameter-update method demonstrates
a reliable aggregate closed-loop gain.
\end{samepage}

\section{Limitations, Safety, and Reproducibility}
\paragraph{Experimental scope.}
The paired CurrentJEPA--FutureJEPA comparison finds no detectable aggregate
future-head gain after 25,000 training steps
(Section~\ref{sec:experiments}). This is not evidence that the effect is
exactly zero: replication with independent training runs remains necessary.
The larger BaseFlow--FutureJEPA difference changes both JEPA conditioning
and future supervision, so it does not identify the future objective's
causal contribution. Shared settings were checked, but exact equality of
all initial parameters and stochastic training draws was not established.

\paragraph{Representation and generalization.}
Future latent MSE can improve without improving action selection.
The small separation between matched and shuffled conditioning illustrates
why prediction metrics must be paired with closed-loop evaluations.
The present evidence is simulation-based and concentrated on LIBERO-Plus.
It does not establish transfer to new embodiments, real cameras, or
long-horizon household tasks.

\paragraph{Context adaptation.}
The complete retry study allows one attempt for the no-context baseline
and up to five attempts for context reuse. Its success increase therefore
combines memory, repeated execution, and extra compute. A complete
five-attempt no-context comparison and a shuffled-context comparison are
still needed to isolate the memory effect; the varying-noise comparison
covers only a subset of cases. Initial pairing checks pass, but some
first-attempt trajectories differ despite zero context, and one later
observation mismatch remains; these uncertainties are reported with the
results. The adapter was trained with an earlier policy and transferred
to the evaluation policy without retraining. Its final action residual
has no hard norm bound. The persistent and transient contexts are intended
to separate environment memory from local correction, but this interpretation
has not been established as an identifiable physical decomposition.

\paragraph{Safety.}
Real-robot use requires workspace and velocity limits, collision monitoring,
an emergency stop, human supervision, and validation of camera calibration
and timing. Simulation success and accepted latent-gradient updates are
not evidence of deployment safety.

\paragraph{Reproducibility.}
Success counts and paired tests are derived from episode-level records.
The authors retain the training configurations and evaluation records.
Independent reproduction additionally requires the benchmark data, visual
and language-model weights, trained policies, and simulator environment,
which are not bundled with the paper. Available checkpoint and source
records do not establish complete historical run-time identity.

\section{Conclusion and Future Work}
We study a future-conditioned extension of a VLM-conditioned DiT flow
policy. The method adds training-only residual prediction heads at offsets
4 and 16 while leaving the inference-time action route unchanged.

The complete paired evaluation yields two conclusions. First, current JEPA
conditioning provides the main success-rate gain: CurrentJEPA reaches
82.45\% versus 71.63\% for BaseFlow under the final mixed-cap protocol.
Second, the future objective does not produce a detectable aggregate
rollout gain at this training budget: FutureJEPA reaches 81.99\% versus
82.45\% for CurrentJEPA under the final mixed-cap protocol, and the strict
paired analysis is a near tie.

Bounded retry with inferred context improves the retry system's success
rate, but it uses a larger attempt budget and therefore should be read as
an operational recovery result rather than a direct estimate of memory's
causal effect. Parameter-level online adaptation likewise has not shown a
reliable aggregate control gain.

The immediate next step is replication with independent training runs at
the same budget, together with matched five-attempt no-context and
shuffled-context retry controls. Future work should also test
action-conditioned latent dynamics, uncertainty-aware replanning,
task-disjoint and cross-embodiment transfer, multimodal grounding, and
real-robot safety validation.

\clearpage
\balance
\small
\setlength{\bibsep}{2pt}
\renewcommand{\bibpreamble}{\interlinepenalty=10000}
\bibliographystyle{unsrtnat}
\bibliography{references}

\clearpage
\nobalance
\appendix
\section{Evaluation Protocol and Additional Retry Studies}
\label{app:protocol-provenance}
\setlength{\intextsep}{8pt plus 2pt minus 2pt}
\setlength{\textfloatsep}{8pt plus 2pt minus 2pt}

\paragraph{Main policy evaluation.}
The primary comparison covers all 10,030 LIBERO-Plus episodes. The final
success-rate tables use 300-step caps on Spatial/Object and 600-step caps
on Goal/LIBERO-10. Each policy uses the same action sampler, initial
states, and paired action-noise streams. The future-prediction heads are
absent from inference.

\begin{table}[!htbp]
\begin{reporttable}[0.94\linewidth]
\caption{Main LIBERO-Plus evaluation settings.}
\label{tab:eval-protocol}
\renewcommand{\arraystretch}{1.04}
\begin{tabular}{@{}>{\raggedright\arraybackslash}p{0.24\linewidth}>{\raggedright\arraybackslash}p{\dimexpr0.76\linewidth-2\tabcolsep\relax}@{}}
\toprule
Component & Setting \\
\midrule
Action chunk & $H \times D_a = 16 \times 32$ native width, cropped to 7 DoF \\
Solver & 4-step uniform Euler; 4 DiT evaluations per chunk \\
Replan interval & 16 environment steps, open-loop chunk execution \\
Episode cap & 300 Spatial/Object; 600 Goal/LIBERO-10 \\
Trials & One trial per task \\
Pairing & Initial state and action-noise stream matched across policies \\
Task counts & Spatial 2,402 / Object 2,518 / Goal 2,591 / LIBERO-10 2,519 \\
Total & 10,030 episodes per policy \\
\bottomrule
\end{tabular}
\reporttablenote{The strict primary paired statistics use the original
300-step cap on every suite.}
\end{reporttable}
\end{table}

\paragraph{Additional retry studies.}
The smaller studies below use selected cases, different policy checkpoints,
or different attempt budgets. Their counts are not pooled with the main
benchmark or with one another.

\begin{table}[!htbp]
\begin{reporttable}[0.94\linewidth]
\caption{Selected retry studies and their scope.}
\label{tab:retry-layers}
\renewcommand{\arraystretch}{1.04}
\begin{tabular}{@{}>{\raggedright\arraybackslash}p{0.18\linewidth}>{\raggedright\arraybackslash}p{0.40\linewidth}>{\raggedright\arraybackslash}p{\dimexpr0.42\linewidth-4\tabcolsep\relax}@{}}
\toprule
Study & Policy, cases, and budget & Result and interpretation \\
\midrule
Complete retry evaluation &
CurrentJEPA policy; all four suites; one baseline attempt and up to five
context-adapted attempts &
No-context/context methods: 8,124/8,914/8,901 successes. The unequal
attempt budget prevents attributing the difference to memory alone. \\
\addlinespace[3pt]
Fixed-noise selected cases &
Earlier FutureJEPA policy; 16 selected records, 14 unique initial states;
up to 20 attempts &
Carry $m$/reset $u$: 14/16; reset both: 0/16. The zero-context first
attempt already succeeds on 14/16. \\
\addlinespace[3pt]
Persistent-context pilot &
Earlier FutureJEPA policy; 26 selected cases; repeated retries &
Memory-only / no-context-only successes: 12/1. Selection and repeated
attempts limit causal interpretation. \\
\addlinespace[3pt]
Spatial subset &
CurrentJEPA policy; 500 Spatial cases; up to five attempts &
Successes: 396 without context, 423 with fixed decay, and 420 with
evidence gating. This is a selected subset. \\
\bottomrule
\end{tabular}
\reporttablenote{Differences in policy, case selection, and retry budget
preclude pooling these results.}
\end{reporttable}
\end{table}

\paragraph{Data and implementation provenance.}
The authors retain the episode-level evaluation records and configuration
files used for the reported analyses. Some original observations were
repaired and re-evaluated. Historical evaluation records do not preserve
all run-time checkpoint identities, so complete exact reproduction of
the original runs cannot be established from those records alone.

\end{document}